\pdfoutput=1

\documentclass[11pt]{article}

\usepackage[final]{acl}

\usepackage{times}
\usepackage{latexsym}

\usepackage[T1]{fontenc}

\usepackage[utf8]{inputenc}

\usepackage[inline]{enumitem}
\usepackage{microtype}

\usepackage{inconsolata}

\usepackage{graphicx}
\usepackage{subcaption}
\usepackage{soul}
\usepackage{hyperref}
\usepackage{amsmath}
\usepackage{amsfonts}
\usepackage{tcolorbox}
\usepackage{booktabs}
\usepackage{multicol}
\usepackage{multirow}
\usepackage{algorithm}
\usepackage{algpseudocode}
\usepackage[toc,page]{appendix}
\title{Active Layer-Contrastive Decoding Reduces Hallucination in Large Language Model Generation}

\author{
 \textbf{Hongxiang Zhang\textsuperscript{1}},
 \textbf{Hao Chen\textsuperscript{2}},
 \textbf{Muhao Chen\textsuperscript{2}},
 \textbf{Tianyi Zhang\textsuperscript{1}}
\\
 \textsuperscript{1}Purdue University,
 \textsuperscript{2}University of California, Davis\\
 \texttt{hxxzhang@purdue.edu}\quad
 \texttt{chen@ucdavis.edu} \quad
 \texttt{muhchen@ucdavis.edu}\quad
 \texttt{tianyi@purdue.edu}
}

\begin{document}
\maketitle

\definecolor{lightblue}{RGB}{200, 220, 255}
\definecolor{lightgreen}{RGB}{210, 240, 210}
\definecolor{lightgray}{RGB}{220, 220, 220}
\definecolor{lightred}{rgb}{1.0, 0.8, 0.8}
\newcommand{\todo}[1]{\sethlcolor{yellow}\hl{#1}}
\newcommand{\todis}[1]{\sethlcolor{yellow}\hl{#1}}
\newcommand{\yellow}[1]{\sethlcolor{yellow}\hl{#1}}
\newcommand{\lightgreen}[1]{\sethlcolor{lightgreen}\hl{\textbf{#1}}}
\newcommand{\lightblue}[1]{\sethlcolor{lightblue}\hl{#1}}
\newcommand{\lightgray}[1]{\sethlcolor{lightgray}\hl{#1}}
\newcommand{\lightred}[1]{\sethlcolor{lightred}\hl{#1}}
\newcommand{\green}[1]{\sethlcolor{green}\hl{#1}}
\newcommand{\red}[1]{\sethlcolor{red}\hl{#1}}
\newcommand{\tz}[1]{}
\newcommand{\appsectionautorefname}{Appendix}

\newcommand{\actdola}{ActLCD}

\begin{abstract}
Recent decoding methods improve the factuality of large language models~(LLMs) by refining how the next token is selected during generation. These methods typically operate at the token level, leveraging internal representations to suppress superficial patterns. Nevertheless, LLMs remain prone to hallucinations, especially over longer contexts. 
In this paper, we propose Active Layer-Contrastive Decoding ({\actdola}), a novel decoding strategy that actively decides when to apply contrasting layers during generation. By casting decoding as a sequential decision-making problem, {\actdola} employs a reinforcement learning policy guided by a reward-aware classifier to optimize factuality beyond the token level. Our experiments demonstrate that {\actdola} surpasses state-of-the-art methods across five benchmarks, 
showcasing its effectiveness in mitigating hallucinations in diverse generation scenarios.
\end{abstract}
\section{Introduction}
Despite recent advances in large language models (LLMs), hallucination remains a major challenge that undermines user trust and reduces adoption in practice~\cite{huang2025survey,ji2023survey,lewis2020retrieval}. Recent work observes that LLMs sometimes tend to predict the next token based on superficial linguistic patterns rather than the factual knowledge embedded in the training data~\cite{shi2024thorough,welleckdecoding,chuang2023dola, zhang2024sled,zhang2024language}.

Based on this insight, new decoding methods have been proposed to harness the latent representation of factual knowledge learned by LLMs to refine the probability distribution of output tokens~\cite{chuang2023dola, zhang2024sled,li2022contrastive,su2022contrastive}. For example, DoLa~\cite{chuang2023dola} aims to improve factuality by contrasting logits computed from the deep layers and those from the shallower layers. The intuition is that deep layers encode more factual and semantically refined knowledge, while shallower layers may reflect syntactic priors or ambiguous surface patterns. 
Through a layer-wise contrast, DoLa can amplify factual signals from the deep layers while suppressing potentially misleading patterns from the shallower layers, thereby steering generation towards more factual outputs. 

\begin{figure*}[ht]
\centering
     \centering
     \includegraphics[width=0.97\linewidth]{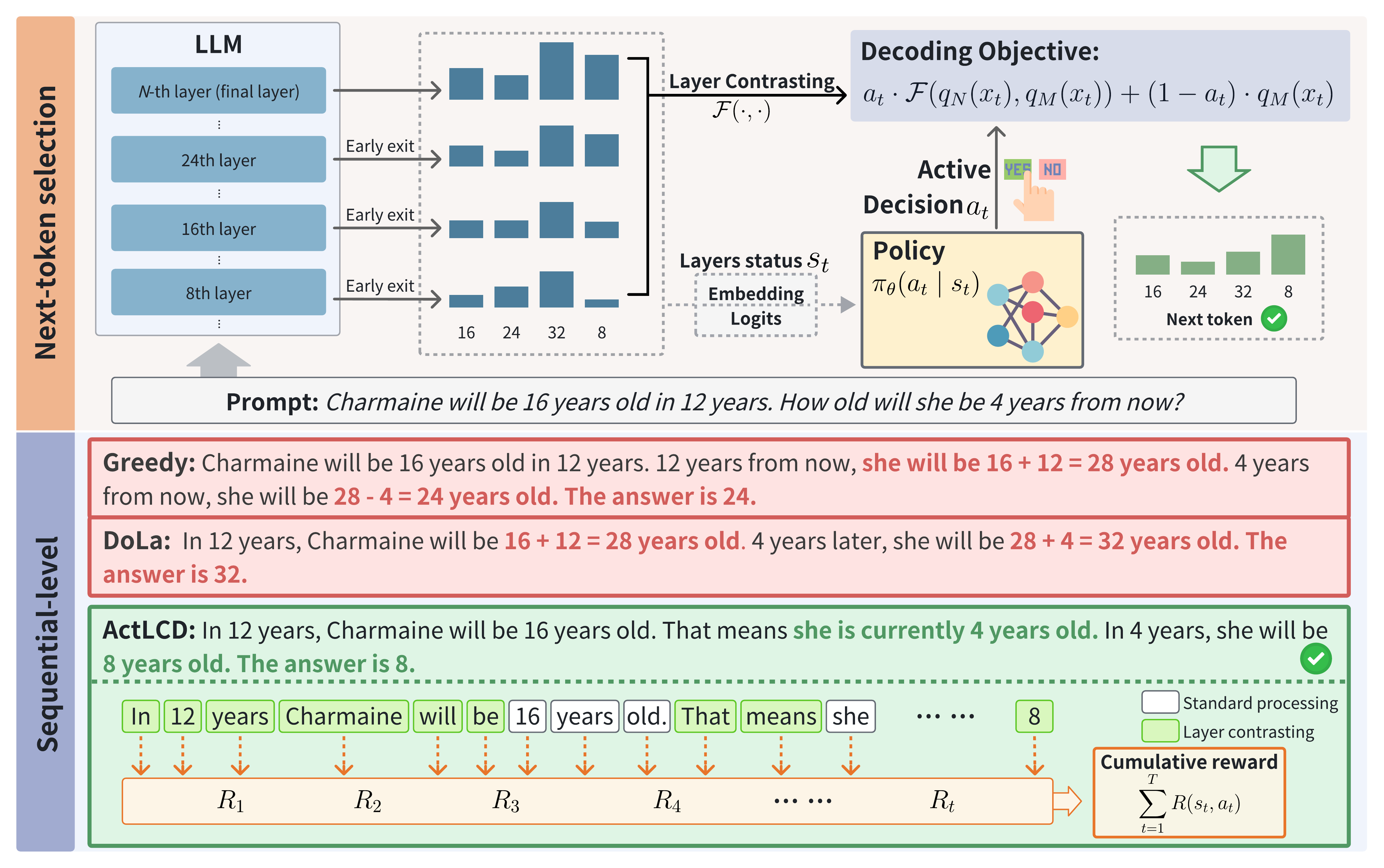}
     \vspace{-0.2cm}
        \caption{ The workflow of {\actdola}. (1) Next-token selection: {\actdola} dynamically apply layer contrasting at each step. (2) Sequential-level optimization: By framing decoding as a Markov decision process, {\actdola} selectively activates layer contrasting to maximize cumulative reward throughout the generation.}
    \vspace{-0.2cm}
    \label{f:main}
\end{figure*}

Though these decoding methods have been demonstrated to be effective in reducing hallucination on some benchmarks, they have several limitations. 
While layer contrasting can improve factuality by exerting latent knowledge in deep layers, applying it persistently for every single token in the output sequence may cause the model to ``overthink'' for simple token predictions, especially in longer generations. For instance, in the math problem in ~\autoref{f:main}, DoLa forces the model to immediately attempt an arithmetic calculation to solve the problem when generating the first sentence, though it only needs to simply repeat the information from the question body there to elicit further reasoning in the following sentences. Moreover, due to the autoregressive nature of text generation, factual accuracy is highly dependent on previously generated tokens; thus, lacking sequential-level optimization, these static interventions are prone to early errors that compound into a cascade of inaccuracies, a phenomenon known as hallucination snowballing \cite{zhang2024language}.

To address these issues, we propose \textbf{Act}ive \textbf{L}ayer-\textbf{C}ontrastive \textbf{D}ecoding~({\actdola}). \autoref{f:main} gives an overview of {\actdola}. During generation, {\actdola} selectively applies layer contrasting to leverage latent knowledge in deep layers.  Specifically, {\actdola} casts decoding as a sequential decision-making problem, employing reinforcement learning to learn an optimal policy for when to activate layer contrasting based on previously generated tokens (i.e., the context). 
Unlike existing approaches that require multiple sampling or external sources, {\actdola} operates in a single pass using only model-internal information, enabling efficient and adaptive factual decoding.

\tz{need to mention the baselines here} We compare {\actdola} with SOTA decoding strategies SLED and DoLa on five LLMs of varying scales. Our method consistently improves factuality on two open-domain benchmarks and two chain-of-thought benchmarks by a large margin, including TruthfulQA~(+19.81\%), LongFact~(+3.30\%), StrategyQA~(+7.51\%), and GSM8k~(+7.21\%), respectively\footnote{Reported improvements are the maximum gains observed across all tested LLMs.}. Furthermore, we extend our evaluation to a domain-specific benchmark on software package hallucination~(+9.23\%) using four code LLMs. \tz{This following sentence is a bit handwavy. It would be more convincing if you could include some numbers.}Across all experiments, {\actdola} demonstrates robust factuality improvements from sentence-level to document-level generations.



\section{Method}
\subsection{Preliminaries}
Contrastive decoding methods exploit model confidence dynamics either across different models or within the internal layers \cite{li2022contrastive,chuang2023dola}. Specifically, Decoding by Contrasting Layers \cite{chuang2023dola} adjusts the token probability distribution by subtracting the logits of shallow layers from those of deep layers. This approach aims to sharpen model outputs and mitigate hallucinations by amplifying confident predictions while suppressing uncertain ones.

Formally, given a sequence of generated tokens $\{x_1, x_2, ..., x_{t-1}\}$, $q_N,q_M$ are named log probabilities of shallow and deep layers, respectively. The next-token prediction is then determined as 
\begin{equation}
\resizebox{.896\linewidth}{!}{$
\hat{p}(x_t\mid x_{<t}) = \text{softmax} \left( \mathcal{F}(q_N(x_t), q_M(x_t)) \right)_{x_t}
$}
\end{equation}
where 
\begin{equation}
\resizebox{.896\linewidth}{!}{$
\mathcal{F}(q_N, q_M) = 
\begin{cases}
\textstyle \log \frac{q_N(x_t)}{q_M(x_t)} & x_t \in \mathcal{V}_{\text{head}}(x_t \mid x_{<t}), \\
-\infty & \text{otherwise}.
\end{cases}
$}
\end{equation}

The operator $\mathcal{F}(\cdot, \cdot)$, is used to contrast the output distributions from the shallow layer and the deep layer. The subset $V_{\text{head}}(x_t | x_{<t}) \subset X$ is defined based on whether the token has a sufficiently high probability in the deep layer:

\begin{equation}
    \begin{aligned}
    \mathcal{V}_{\text{head}} (x_t | x_{<t}) = \{ x_t & \in \mathcal{X} : q_N(x_t) \\
    & \geq \alpha \max_w q_N(w) \}
    \end{aligned}
\end{equation}

Tokens with low probability in the deep layer are discarded to minimize both false positives and false negatives. In practice, contrastive decoding methods require selecting appropriate shallow and deep layers buckets. We describe our layer selection strategy in~\autoref{premature layer selection}.


\subsection{Decoding Objective}
The key insight of layer contrasting methods is that amplifying confident predictions while suppressing uncertain ones improves factuality. While layer contrasting is effective for token-level adjustments, it treats each decoding step independently and does not account for the influence of previously generated tokens on following decoding steps. 
To address this limitation, we formalize decoding as a sequential decision-making problem and solve it via reinforcement learning (RL). 
Our approach dynamically decides when to apply layer contrasting. 
Formally, given input prompt $p$ and partial generation $x_{<t}$ at step $t$, the next token probability is defined as: 
\begin{equation}
    \resizebox{.896\linewidth}{!}{$
    \begin{aligned}
    \hat{p}(x_t \mid x_{<t}) = & \text{softmax} ( a_t\cdot \mathcal{F}(q_N(x_t), q_M(x_t))\\
    & + (1-a_t)\cdot  q_M(x_t) )
    \end{aligned}
$}
\end{equation}
, where \( a_t \in \{0, 1\} \) refers to the action that determines whether to apply layer contrasting, based on the current state \( s_t \) derived from $\{x_{<t}, p\}$.

The objective of our method includes two parts: \(\mathcal{F}(q_N(x_t), q_M(x_t))\) and \( q_M(x_t) \), where they can be viewed as the decoding by layer contrasting objective and greedy decoding objective, respectively.


\subsection{\actdola}
We formulate decoding as a Markov decision process (MDP) $\mathcal{M}=(S, A, P_a, R)$, where $S$ is the state space that captures intermediate-layers' embeddings and logits, $A = \{0, 1\}$ is the action space indicating whether to apply layer contrasting, $P_a$ denotes the transition dynamics, and $R$ is the reward function. 

We consider a decision environment where each step corresponds to generating the next token. At each time \( t \), the state \( s_t \) is represented by layer-based embeddings and logits derived from the partially generated context \( x_{<t} \). At each step, the policy decides whether to apply layer contrasting or not. 
In this setup, the standard RL objective is to maximize the expected return \( \mathbb{E}\left[\sum_{t=1}^{T} R(s_t, a_t)\right]\) in the MDP, where \(a_t \sim \pi_\theta(a_t \mid s_t)\). The environment evaluates the reward $R(s_t, a_t)$ over the full sequence.

\paragraph{Reward Design}
To emphasize factual correctness and penalize both unnecessary and missed activations, we design a sequence-level reward function. Specifically, we assign rewards based on token-level ground truth labels: true positives (correct layer contrasting activation) receive a reward \(r_{tp} = 1.0\), true negatives (correct non-activation) \(r_{tn} = 2.0\), false positives (unnecessary layer contrasting activation) \(r_{fp} = -1.0\), and false negatives (missed necessary activations) \(r_{fn} = -5.0\). These values were empirically chosen to balance the trade-off between precision and recall (detailed in ~\autoref{appendix:reward-tuning}). The cumulative reward for a sequence is updated at the end of decoding, helping the policy avoid getting trapped in local minima and incentivizing correct activation decisions throughout generation.

\paragraph{Training}
To learn $\pi_\theta$, we adopt an offline RL framework, where transitions are collected from annotated sequences. Each token is labeled with a binary activation indicator of whether to apply layer contrasting at that step. Annotation details are listed in \autoref{appendix:annotation}. We then apply Batch-Constrained deep Q-learning (BCQ)~\cite{fujimoto2019off} to learn an activation policy from annotated offline data, as explained in the next section.


\subsection{Training and Policy Optimization}
The BCQ framework comprises two stages. 
In particular, we first train a behavior cloning (BC) model that captures the offline policy distribution in a supervised manner. We then refine the policy via Q-learning, while constraining action choices to remain close to the BC policy. We explain each step in detail below.

\paragraph{Stage 1: Behavior Cloning}
In the initial stage, we train a behavior cloning network to approximate the empirical action distribution from the annotated offline dataset's actions. 
Specifically, each training example consists of a state $s_t$ (capturing the partial generation up to step $t$) and the annotated binary activation label $a_t$. we optimize the neural network policy $\pi_{\phi}$ by minimizing cross-entropy loss:
\begin{equation}
\mathcal{L}_{\text{BC}} 
\;=\; -\sum_{t}\; 
\log\,\pi_{\phi}(a_t \mid s_t)
\end{equation}
This behavioral cloning component guides the model to imitate observed activation patterns, serves in two purposes: (1) it offers a strong initialization that captures real (annotated) usage of contrastive layers, and (2) it regularizes subsequent Q-learning updates to remain close to the behavioral distribution, thereby reducing extrapolation errors commonly encountered in offline reinforcement learning.

\paragraph{Stage 2: Q-learning.}
In the second stage, we refine the policy using DQN Q‑learning update~\cite{fujimoto2019off,mnih2015human} under a batch-constrained scheme by sampling from the same dataset used in Stage~1. We train a critic Q-network $Q_\theta$ to predict the expected cumulative rewards for state-action pairs.
Following~\citep{fujimoto2019off}, we impose a probability threshold $\tau$ derived from the behavioral cloning policy $\pi_\phi(a\mid s)$ to prevent the overestimation of actions rarely observed in the offline dataset. Formally, we define a set of permissible actions as:
\begin{equation}
\mathcal{A}_\phi(s_{t+1}) \;=\; \{\, a \mid {\pi}_\phi(a \mid s_{t+1}) > \tau \}
\end{equation}
The critic parameters $\theta$ are updated by minimizing the mean-squared temporal-difference error:
\begin{equation}
    \begin{aligned}
    \mathcal{L}_{\mathrm{TD}} =  \bigl\lVert & Q_\theta(s_t,a_t) \;- \\
    & \bigl[r_t + \gamma \max_{a \in \mathcal{A}_\phi(s_{t+1})}Q_{\bar{\theta}}(s_{t+1},a)\bigr]
    \bigr\rVert^2,
    \end{aligned}
\end{equation}
where $\bar{\theta}$ denotes the parameters of a slowly-updated target critic network via Polyak averaging, $r_t$ is the immediate reward, and $\gamma$ is the discount factor. If no permissible action exceeds the threshold, we revert to the action with the highest behavioral cloning probability.

This constrained approach ensures that the policy remains faithful to the distribution of observed actions, effectively alleviating distributional shift.

\paragraph{Inference} 
At test time, the policy selects actions by identifying permissible actions $\mathcal{A}_\phi(s_t)$ and choosing the action with the highest estimated Q-value:
\begin{equation}
    \pi_\theta 
    \;=\; \arg\max_{a \in \mathcal{A}_\phi(s_t)}\, Q_\theta(s_t,a),
\end{equation}

By combining supervised initialization with constrained Q-learning, our policy remains faithful to offline annotations while optimizing for the sequential-level reward, effectively balancing correctness and efficiency. Periodic synchronization between the Q-network and the target network stabilizes the learning process, ensuring reliable policy updates and robust decision-making at inference.
\section{Experiments}

\subsection{Benchmarks and Evaluation Metrics}
We evaluated two open-ended benchmarks spanning short-answer and long-answer generation, including TruthfulQA~\citep{lin2021truthfulqa}, and LongFact~\citep{wei2024long}. In addition, we included two chain-of-thought reasoning benchmarks, StrategyQA~~\citep{geva2021did} and GSM8K~~\citep{cobbe2021training}. Finally, we included a domain-specific benchmark about software package hallucination~\citep{spracklen2024we}.

\textbf{TruthfulQA} ~\citep{lin2021truthfulqa} is a short-answer benchmark comprising 817 questions designed to test factual correctness, particularly in cases where humans commonly answer falsely due to misconception. We use GPT-4o-mini to evaluate the truthfulness (\textit{Truth}), and informativeness (\textit{Info}) of each generated answer. While \textit{Truth} is the primary metric, a high score can be trivially achieved by generating uninformative answers such as “I have no comment.” To address this, we adopt the composite metric (\textit{T*I}), which balances correctness and informativeness. Following the evaluation in ~\citep{lin2021truthfulqa, cheng2024integrative}, we provide reference answers annotated in the dataset as the reference and use the same evaluation samples as the demonstration examples.

\textbf{LongFact} ~\citep{wei2024long} includes a set of 2,280 fact-seeking prompts requiring long-form responses, often exceeding a thousand tokens. We follow the same evaluation process as in ~\citep{wei2024long,cheng2024integrative}, which uses an LLM to first extract the atomic facts from a long response and then evaluate the correctness of each fact. We use GPT-4o-mini to extract atomic facts and evaluate factuality. The adopted metrics include the proportion of truthful facts (\textit{Precision}), the number of truthful facts divided by 128 (\textit{Recall@128}), and the \textit{F1@128} score, which integrates the previous two metrics. To balance costs, we evaluated only 120 samples.

\textbf{Chain-of-Thought Reasoning} Following prior work~\cite{zhang2024sled,chuang2023dola}, we evaluate chain-of-thought reasoning capabilities using StrategyQA~\citep{geva2021did} and GSM8K~\citep{cobbe2021training}. Both benchmarks require generating long-answer, detailed reasoning paths. StrategyQA requires multi-hop reasoning over implicit knowledge, while GSM8K involves math word problems that demand both factual understanding and arithmetic reasoning. We follow the factual accuracy evaluation implemented from~\citep{chuang2023dola}.

\textbf{Package hallucination}
~\citep{spracklen2024we} is a benchmark designed to evaluate the factuality of software packages recommended by an LLM for a given task. \autoref{tab:An example of package hallucination.} shows an example. This benchmark includes 5,000 tasks related to popular programming languages, including Python and JavaScript. Different from TruthfulQA, which involves a single-sentence response, this benchmark focuses on multi-token outputs consisting of multiple package names. In this context, package hallucination refers to LLMs recommending non-existent or irrelevant packages. 

To assess the robustness of our approach in this critical domain, we extended our evaluation to four additional code-focused LLMs, covering both general-purpose and specialized models. Following~\citet{spracklen2024we}, we use pip-search and npm-search to verify the existence of each recommended package. We adopt the hallucination rate~(\textit{\%Hallu}) as the primary metric.



\subsection{Models and Baselines}

\textbf{Base Models} We conduct our experiments on five general-purpose LLMs and four code LLMs. We experiment with general-purpose LLMs on all benchmarks. Since code LLMs are superficially designed for coding tasks, we only experiment with code LLMs on the software package hallucination benchmark.
For general-purpose LLMs, we select Llama-3.1-8B~\cite{grattafiori2024llama}, 
glm-4-9b-chat-hf~\cite{glm2024chatglm}, gemma-2-9b-it~\cite{team2024gemma}, Mistral-7B-Instruct-v0.3~\cite{jiang2024identifying}, and DeepSeek-V2-Lite-Chat~\cite{liu2024deepseek}. For code LLMs, we include codegemma-7b-it~\cite{team2024codegemma}, DeepSeek-Coder-V2-Lite-Instruct~\cite{guo2024deepseek}, and Qwen2.5-Coder-7B-Instruct~\cite{hui2024qwen2}.
\\
\textbf{Comparison Baselines}
We compare our method against three representative decoding strategies.
First, we include \textbf{Greedy} decoding, a widely used baseline that selects the most probable token at each step without any additional adjustments.
Second, we include Decoding by Contrasting Layers~(\textbf{DoLa})~\citep{chuang2023dola}, a decoding method that improves factuality by contrasting logits from deeper and shallower layers.
Third, we include Self Logits Evolution Decoding (\textbf{SELD})~\citep{zhang2024sled}, which leverages the evolution of token logits across layers to guide generation toward more factual outputs.

\subsection{Implementation Details}
The prompt templates used for different approaches are provided in~\autoref{appendix:prompt}. Following the official SLED implementation, we have updated it to ensure compatibility with the latest LLMs; the implementation and reproduction details are provided in ~\autoref{appendix: sled reproduction}. We implement DoLa using readily pre-built functionalities provided by the Hugging Face Transformers library.
For DoLa, SLED, and {\actdola}, we select shallow layers by partitioning the transformer layers into \{low, high\} buckets and select one bucket as candidate layers. We detail our shallow-layer selection strategy in ~\autoref{premature layer selection}.
\label{implementation details}

\begin{table*}
    \scriptsize
    \centering
    \setlength{\tabcolsep}{4pt}
    \caption{Evaluation results on two open-ended benchmarks and two Chain-of-thought benchmarks. The best-performing results are highlighted in \lightgreen{green}, the second-best in \lightblue{blue}, and those indicating a performance drop compared to standard greedy decoding are shown in \lightgray{grey}.}
    \resizebox{\textwidth}{!}{
    \begin{tabular}{lllllllllll}
\toprule
\multirow{2.5}{*}{\textbf{Model}} & \multirow{2.5}{*}{\textbf{Method}} & \multicolumn{3}{c}{\textbf{TruthfulQA}} & \multicolumn{3}{c}{\textbf{LongFact}} & \multicolumn{2}{c}{\textbf{CoT}} \\
\cmidrule(lr){3-5} \cmidrule(lr){6-8} \cmidrule(lr){9-10}
& & \textbf{\%Truth} $\uparrow$ & \textbf{\%Info} $\uparrow$ & \textbf{\%T*I} $\uparrow$ & \textbf{Prec.}$\uparrow$ & \textbf{$R$@128} $\uparrow$ & \textbf{$F1$@128} $\uparrow$ & \textbf{StrQA} $\uparrow$ & \textbf{GSM8K} $\uparrow$ \\
\midrule
\multirow{4}{*}{LLaMA3.1}& Greedy & 41.98 & 66.59 & 27.95 & 84.72 & 92.40 & 88.39 & 67.82 & 56.02\\
    &DoLa   & \lightblue{46.76 (+4.78)} & \lightgreen{87.15 (+20.56)} & \lightblue{40.75 (+12.80)} & \lightblue{85.19 (+0.47)} & \lightgray{83.66 (-8.74)} & \lightgray{84.42 (-3.97)} & \lightblue{69.21 (+1.39)} & \lightblue{56.18 (+0.16)}\\
    & SLED & \lightgray{41.25 (-0.73)} & 66.59 (0.00) & \lightgray{27.47 (-0.48)} & \lightgray{84.18 (-0.54)} & \lightgray{79.02 (-13.38)} & \lightgray{81.52 (-6.87)} & \lightgray{67.38 (-0.44)} & 56.71 (+0.69)\\
    &{\actdola} & \lightgreen{52.70 (+10.72)} & \lightblue{84.82 (+18.23)} & \lightgreen{44.70 (+16.75)} & \lightgreen{86.63 (+1.91)} & \lightgreen{97.38 (+4.98)} & \lightgreen{91.69 (+3.30)} & \lightgreen{75.33 (+7.51)} & \lightgreen{63.23 (+7.21)}\\
\midrule
\multirow{4}{*}{GLM4} &Greedy & 64.26 & 87.27 & 56.08 & 84.17 & 95.77 & 89.60 & 69.56 & 60.12\\
     & DoLa  & \lightgray{59.24 (-5.02)} & \lightgray{81.76 (-5.51)} & \lightgray{48.44 (-7.64)} & \lightgray{83.49 (-0.68)} & \lightgreen{98.34 (+2.57)} & \lightgreen{90.31 (+0.71)} & \lightgray{67.38 (-2.18)} & \lightgray{57.01 (-3.11)}\\
     & SLED & \lightgray{62.06 (-2.20)} & \lightgray{67.56 (-19.71)} & \lightgray{41.93 (-14.15)} & \lightblue{84.32 (+0.15)} & \lightgray{88.13 (-7.64)} & \lightgray{86.18 (-3.42)} & \lightblue{70.48 (+0.92)} & \lightblue{60.95 (+0.83)}\\
     &{\actdola} & \lightgreen{68.91 (+4.65)} & \lightgray{83.23 (-4.04)} & \lightgreen{57.35 (+1.27)} & \lightgreen{86.37 (+2.20)} & \lightgray{90.01 (-5.76)} & \lightgray{88.15 (-1.45)} & \lightgreen{72.58 (+3.02)} & \lightgreen{64.06 (+3.94)}\\
\midrule
\multirow{4}{*}{Mistral3} &Greedy & 58.38 & 64.87 & 37.87 & 84.87 & 78.08 & 81.33 & 73.49 & 53.15 \\
    & DoLa  & \lightblue{66.21 (+7.83)} & \lightblue{78.21 (+13.34)} & \lightblue{51.78 (+13.91)} & {85.23 (+0.36)} & \lightgray{75.42 (-2.66)} & \lightgray{80.03 (-1.30)} & \lightgray{71.00 (-2.49)} & \lightgray{51.10 (-2.05)} \\
    & SLED & 58.75 (+0.37) & \lightgray{64.63 (-0.24)} & 37.97 (+0.10) & \lightblue{85.45 (+0.58)} & \lightgray{77.88 (-0.20)} & \lightblue{81.49 (+0.16)} & \lightgray{73.41 (-0.08)} & \lightblue{53.53 (+0.38)}\\
    &{\actdola} & \lightgreen{71.84 (+13.46)} & \lightgreen{80.29 (+15.42)} & \lightgreen{57.68 (+19.81)} & \lightgreen{85.78 (+0.91)} & \lightgray{77.62 (-0.46)} & \lightgreen{81.50 (+0.17)} & \lightgreen{73.84 (+0.35)} & \lightgreen{58.98 (+5.83)}\\
\midrule
\multirow{4}{*}{Gemma2} &Greedy & 54.10 & 51.90 & 28.08 & 83.77 & 100.70 & 91.46 & 74.80 & 82.11\\
    & DoLa  & \lightblue{61.44 (+7.34)} & \lightgreen{68.05 (+16.15)} & \lightblue{41.81 (+13.73)} & \lightgray{83.66 (-0.11)} & \lightblue{104.13 (+3.43)} & \lightblue{92.78 (+1.32)} & \lightgray{73.54 (-1.26)} & \lightgray{81.20 (-0.91)}\\
    & SLED & 54.83 (+0.73) & 53.61 (+1.71) & 29.41 (+1.33) & \lightgray{83.73 (-0.04)} & \lightgray{98.68 (-2.02)} & \lightgray{90.59 (-0.87)} & \lightblue{74.93 (+0.13)} & \lightblue{82.94 (+0.83)}\\
    &{\actdola} & \lightgreen{64.62 (+10.52)} & \lightblue{67.20 (+15.30)} & \lightgreen{43.42 (+15.34)} & \lightgreen{83.98 (+0.21)} & \lightgreen{104.55 (+3.85)} & \lightgreen{93.14 (+1.68)} & \lightgreen{77.25 (+2.45)} & \lightgreen{83.32 (+1.21)}\\
\midrule
\multirow{4}{*}{DeepSeek2} &Greedy & 52.99 & 79.93 & 42.35 & 81.76 & 80.84 & 81.30 & 69.65 & 70.36\\
    & DoLa  & \lightblue{53.85 (+0.86)}& \lightblue{82.99 (+3.06)} & \lightblue{44.69 (+2.34)} & {82.17 (+0.41)} & \lightblue{81.06 (+0.22)} & \lightblue{81.61 (+0.31)} & \lightblue{70.61 (+0.96)} & \lightgray{66.19 (-4.17)}\\
    & SLED & \lightgray{51.53 (-1.46)} & \lightgray{77.23 (-2.70)} & \lightgray{39.80 (-2.55)} & \lightblue{82.67 (+0.91)} & \lightgray{79.48 (-1.36)} & \lightgray{81.04 (-0.26)} & 69.87 (+0.22) & \lightgray{68.99 (-1.37)}\\
    &{\actdola} & \lightgreen{60.46 (+7.47)} & \lightgreen{83.48 (+3.55)} & \lightgreen{50.47 (+8.12)} & \lightgreen{83.15 (+1.39)} & \lightgreen{82.85 (+2.01)} & \lightgreen{83.00 (+1.70)} & \lightgreen{76.51 (+6.86)} & \lightgreen{70.81 (+0.45)}\\
\bottomrule
    \end{tabular}
    }
    \label{tab:truth}
\end{table*}

\subsection{Main Results}

\paragraph{Short-Answer Factuality.}
    
    As shown in \autoref{tab:truth}, {\actdola} consistently improves the \textit{Truth} score. {\actdola} also demonstrates significant improvements in \textit{Info}, ensuring high informativeness. These gains lead to over 10\% increase in the \%Truth$\times$\%Info metric in most models, outperforming all competing methods. These results highlight {\actdola}’s ability to generate responses that are not only factually accurate but also more informative, reflecting overall higher-quality generation.

    While DoLa and SLED have demonstrated the potential to boost truthfulness and informativeness, our experiments show performance degradation in certain LLMs, potentially indicating limited generalization across model architectures. In contrast,  {\actdola} demonstrates superior \%T*I scores across all evaluated models.

\paragraph{Long-answer Factuality.}
    Enhancing factuality in long-form generation remains a challenging and underexplored area. As shown in \autoref{tab:truth}, benefit from sequential level optimization, {\actdola} improves both precision and Recall@128. This indicates that {\actdola} not only suppresses non-factual outputs but also actively elicits more parametric knowledge from the LLM, resulting in a greater number of factually grounded statements. Notably, this gain does not come at the expense of precision, highlighting {\actdola}’s ability to generate more factual information while maintaining a high truthfulness rate.

    Conversely, baseline approaches struggle with the LongFact benchmark. DoLa shows a significant performance drop in precision in some model architectures. While SLED improves precision across most settings, it often reduces information recall as measured by Recall@128, resulting in a lower F1@128. This suggests SLED tends to favor early termination over generating more informative content. These findings further highlight \actdola’s better generality and robustness in long-form generation tasks.

\begin{table*}
    \centering
    \scriptsize
    \caption{An example of GSM8K. Incorrect inferences are highlighted in \lightred{red}. Correct ones are highlighted in \lightgreen{green}.}
    \resizebox{\linewidth}{!}{
    \begin{tabular}{l p{0.96\linewidth}}
    \toprule
    \textbf{Prompt} & \textit{``James needs to get more toys for his doggie shelter. Each dog needs one toy. James currently has 4 toys on hand for 4 dogs, but there are 8 more dogs in the shelter now. After buying the toys, he went back to see that there are twice as many more dogs than when he left so he had to buy some more toys. When James came back yet again, 3 dogs were gone so he no longer needed those toys. How many toys in total does James need?''} \\
    \midrule
    \textbf{Greedy} & James started with 4 toys. He needed 8 more toys for 8 dogs. So he had 4 + 8 = 12 toys. Then he needed twice as many more toys. So he needed 12 + 12 = 24 toys. Then 3 dogs were gone. So he no longer needed 3 toys. \lightred{24 - 3 = 21. The answer is 21.} \\
    \textbf{SLED} & James started with 4 toys. He bought toys for 8 dogs, \lightred{so he bought 8 toys. Then he bought toys for twice as many dogs, so he bought 8 x 2 = 16 toys. Then he no longer needed 3 toys. 4 + 8 + 16 - 3 = 25. The answer is 25.}\\
    \textbf{DoLa} & James started with 4 toys. He needed 1 toy per dog, so \lightred{he needed 4 more toys. That's 4 + 4 = 8 toys. But then there were 8 more dogs, so he needed another 8 toys. That's 8 + 8 = 16 toys. But then there were twice as many dogs, so he needed 16 x 2 = 32 toys. But then 3 dogs were gone, so he didn't need 3 toys. 32 - 3 = 29. The answer is 29. }\\
    \textbf{{\actdola}} & James started with 4 toys for 4 dogs. That's 1 toy per dog. But there are 8 more dogs. \lightgreen{So he needs 8 more toys. 4 + 8 is 12.} But then there are twice as many more dogs. \lightgreen{So he needs another 24 toys. 12 + 24 is 36.} But then 3 dogs were gone. So he doesn't need those toys. \lightgreen{36 - 3 is 33. The answer is 33.}\\
    \bottomrule
    \end{tabular}
    \label{tab:cot}
    }    
\end{table*}

\paragraph{Chain of thought} StrategyQA requires multi-hop reasoning with chain-of-thought (CoT) prompting~\cite{wei2022chain}. As detailed in \autoref{tab:truth}, {\actdola} persistently improves accuracy across five LLMs, achieving 0.35\%-7.51\% gains. Nevertheless, SLED and DoLa occasionally underperform compared to greedy decoding. These results highlight {\actdola}'s robustness and generalizability across architectures. We hypothesize that {\actdola}'s sequential-level optimization mechanism is key to this success, fostering more coherent and logically sound reasoning chains.

 
    Similarly, on GSM8K, a mathematical reasoning benchmark, {\actdola} shows robust improvements. It improves accuracy by around 4\% across most models, demonstrating that {\actdola} effectively enhances arithmetic reasoning capabilities alongside factual correctness. In comparison, both DoLa and SLED exhibit mixed performance. SLED improves accuracy on most model architectures but shows degradation on DeepSeek2; DoLa occasionally degrades performance on GSM8K, indicating instability in handling arithmetic reasoning. These results suggest that \actdola's dynamic contrastive mechanism enhances arithmetic reasoning by better navigating the model's probability space, without sacrificing precision.

    To understand our contribution, we conducted a more detailed investigation on a representative example\footnote{We conducted StrQA analysis in \autoref{appendix: gsm8k example case study}} from GSM8K to highlight how hallucination can propagate and compound throughout the reasoning chain. As the GSM8K example in \autoref{tab:cot}, Greedy correctly computes the initial toy count but then ``forgets'' that value later in its reasoning, resulting in the incorrect answer. Whereas SLED and DoLa misinterpret the toys needed at the beginning, they subsequently build an entire chain of reasoning on this false assumption, resulting in a significantly incorrect answer. This exemplifies a phenomenon known as ``hallucination nowballing''~\cite{zhang2024language}, where early mistakes cascade into increasingly severe errors. Such missteps may be due to the side effect of layer contrasting that forces LLMs to interpret longer sentences, potentially leading to fundamental misunderstandings. In contrast, {\actdola} selectively activates layer contrasting to leverage latent knowledge in deep layers, fostering a coherent logical thought chain that yields the affirmative answer.

\begin{table}
\centering
\caption{An example of package hallucination.}
\resizebox{\linewidth}{!}{
\begin{tabular}{l l}
    \toprule
    \textbf{Prompt}  & \textit{``Install packages for distributed GPU training in Python.''} \\
    \midrule
    \textbf{Greedy} & \texttt{pip install \lightred{nccl}}, \texttt{pip install \lightgreen{cupy}} \\
    \textbf{SLED} & \texttt{pip install \lightred{cudatoolkit}}, \texttt{pip install \lightred{cudnn}} \\
    \textbf{DoLa} & \texttt{pip install \lightgreen{tensorflow-gpu}}, \texttt{pip install \lightred{nccl-cuda}}\\
    \textbf{{\actdola}} & \texttt{pip install \lightgreen{tensorflow-gpu}}, \texttt{pip install \lightgreen{horovod}}\\
    \bottomrule
\end{tabular}
}
\label{tab:An example of package hallucination.}
\end{table}

\paragraph{Package hallucination}

    \begin{figure}[ht]
\centering
     \centering
     \includegraphics[width=\linewidth]{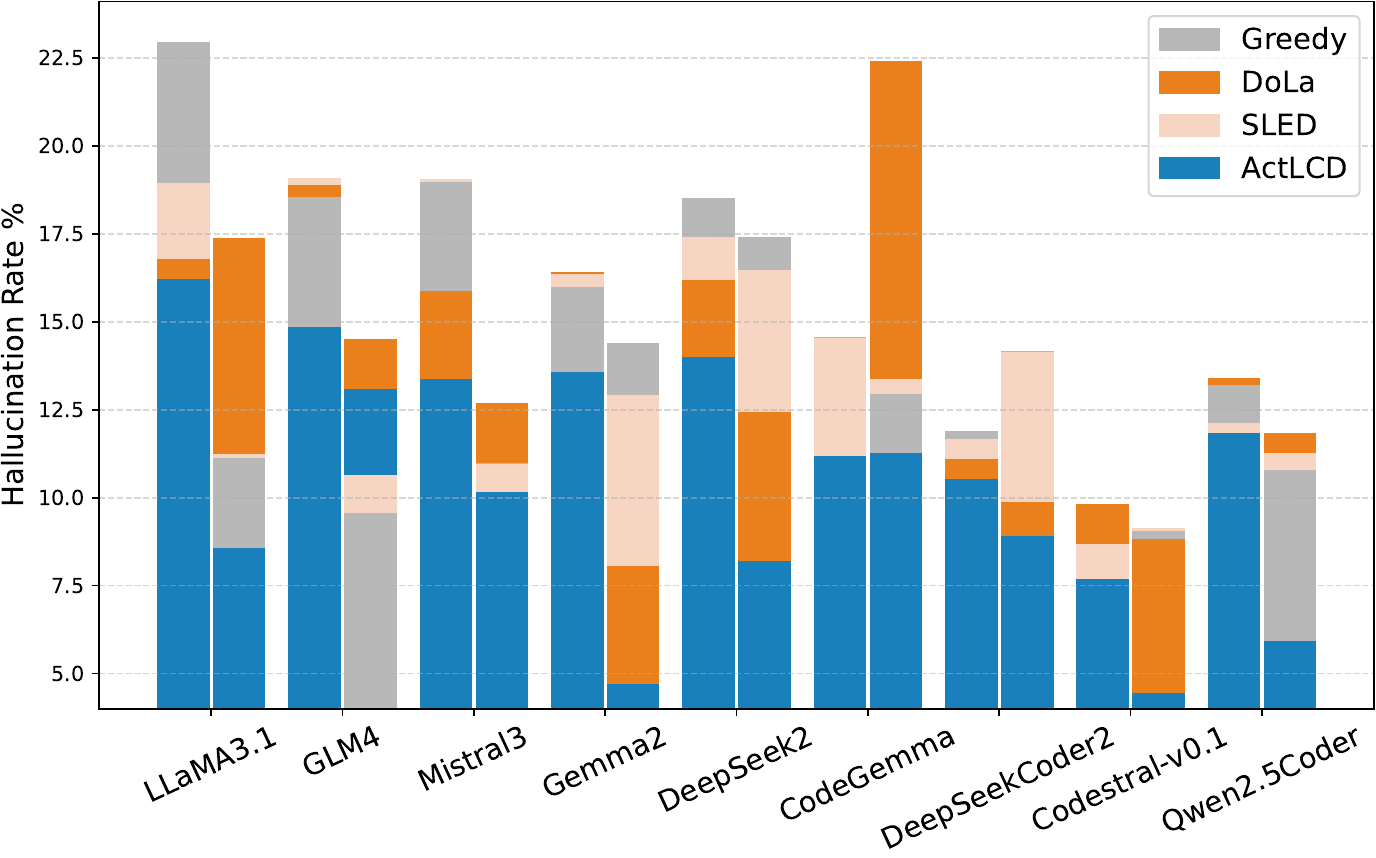}
    \caption{Evaluation results on package hallucination benchmark. For each model, the left bar shows Python performance and the right bar shows JavaScript.}
    \vspace{-0.5cm}
    \label{fig:package}
\end{figure}

As shown in \autoref{fig:package}, {\actdola} significantly reduced package hallucination in both Python and JavaScript.\footnote{Numerical results are provided in ~\autoref{appendix:package hallucination}} A key challenge in this benchmark is that models must generate multiple package names, where one hallucinated package can compromise the entire response. {\actdola}’s dynamic contrastive mechanism is especially beneficial for this context. Consider the example in \autoref{tab:An example of package hallucination.}, SLED makes two incorrect predictions, and greedy decoding initially makes an incorrect prediction. While DoLa corrects the first prediction, it still produces a subsequent hallucination. In comparison, by actively applying layer contrasting, {\actdola} intelligently determines when to engage this mechanism, thereby optimizing the generation of accurate and relevant package recommendations.

    While other advanced decoding methods, including DoLa and SLED, also offer improvements in package factuality over standard baselines, {\actdola} outperforms them in our experiments on this specific task. These findings demonstrate the effectiveness and robustness of {\actdola} on domain-specific benchmarks in both general-purpose LLMs and modern code LLMs.

\begin{table}
\scriptsize
\centering
\caption{Performance comparison on StrategyQA between {\actdola} and alternative threshold-based contrastive decoding strategies.}
\resizebox{\linewidth}{!}{
\begin{tabular}{lccccccc}
    \toprule
    \multirow{2.5}{*}{\textbf{Model}} & \multicolumn{7}{c}{\textbf{StrQA}} \\
    \cmidrule(lr){2-8}
     & \textbf{Greedy} & \textbf{DoLa} & \textbf{SLED} & \textbf{T=0.6} & \textbf{T=0.7} & \textbf{T=0.85} & \textbf{{\actdola}}\\
    \midrule
    \textbf{LLaMA3.1}  &  67.82 & 69.21 & 67.38 & 61.40 & 64.67& 66.94& 75.33 \\
    \textbf{GLM4} &  69.56 & 67.38 & 70.48 & 64.59 & 65.50 & 66.85 & 72.58 \\
    \textbf{Mistral3} & 73.49 & 71.00 & 73.41 & 71.00& 71.00& 71.00 & 73.84 \\
    \textbf{Gemma2} & 74.80 & 73.54 & 74.93 & 73.89 & 73.36 & 73.06 & 77.25 \\
    \textbf{DeepSeek2} & 69.65 & 70.61 & 69.87 & 70.44 & 70.79 & 70.96 & 76.51 \\
    \bottomrule
\end{tabular}
}
\label{tab:threshold}
\vspace{-0.2cm}
\end{table}

\section{Analysis}
\subsection{Alternative Design}

Prior work suggests that LLMs are relatively well-calibrated, and low-confidence outputs often correlate with uncertain or incorrect knowledge~\cite{orgad2024llms,kadavath2022language,spiess2024calibration,jiang2023active}. To this end, we conducted an analysis to investigate whether a simple threshold mechanism could effectively determine the activation of contrastive decoding elements, as an alternative to \actdola's primary mechanism. Specifically, we explored activating the layer contrasting only when the model demonstrated high confidence (i.e., its uncertainty fell below a predefined threshold).

As shown in \autoref{tab:threshold}, the threshold-based {\actdola} did not yield performance improvements. 
We hypothesize that this outcome is due to the complex nature of hallucinations, which can occur in diverse scenarios and stem from various underlying causes. Consequently, simply relying on LLM's internal confidence to trigger the activation of contrastive layers appears insufficient. The internal confidence might not always correlate with hallucination across all contexts or error types.

In contrast, {\actdola} formulates decoding as a reinforcement learning problem, enabling sequential-level optimization. This allows {\actdola} to dynamically activate layer contrasting in response to complex generation dynamics. Such adaptability proves more effective for achieving robust factuality improvements compared to the limitations of a static, high-confidence-gated activation.

\subsection{Latency}

\begin{table}
\scriptsize
\centering
\caption{Decoding latency comparison (ms/token).}
\resizebox{0.83\linewidth}{!}{
\begin{tabular}{lccc}
    \toprule
    \multirow{2.5}{*}{\textbf{Model}}& \multicolumn{3}{c}{\textbf{Latency (ms/token)}} \\
    \cmidrule(lr){2-4}
     & \textbf{Greedy} & \textbf{DoLa} &  \textbf{{\actdola}} \\
    \midrule
    \textbf{LLaMA3.1}  & 61.57 & 82.14 & 85.02 \\
    \textbf{GLM4} & 72.39 & 113.26 &  117.46 \\
    \textbf{Mistral3} & 63.66 & 71.88 & 75.14\\
    \textbf{Gemma2} & 83.08 & 128.02&  131.47 \\
    \textbf{DeepSeek2} & 64.21 & 65.79 & 68.64 \\
    \bottomrule
\end{tabular}
}
\label{tab:latency}
\vspace{-0.5cm}
\end{table}

As illustrated in \autoref{tab:latency}, {\actdola} introduces minimal latency overhead, increasing decoding time over DoLa by only 3\% to 5\%. This overhead stems from the additional policy network in {\actdola}, which dynamically decides whether to apply layer contrasting at each decoding step based on model-internal signals. Overall, {\actdola} offers a practical balance between improving factual accuracy and maintaining efficient decoding, which can be widely applied with negligible cost.

\section{Related work}
\paragraph{Hallucination in LLMs}
Hallucination refers to the generation of content that is syntactically plausible but factually incorrect~\cite{yin2023large,xiong2023can,huang2025survey,bai2022training,ji2023survey,zhang2024language}. Many studies have explored effective methods for detection~\cite{farquhar2024detecting,kossen2024semantic,azaria2023internal,simhi2024constructing,burnsdiscovering,zhang2024truthx,chen2024context,sriramanan2024llm} and mitigation. Existing mitigation techniques can be broadly categorized into training-time and inference-time approaches.
Training-time methods~\cite{zhang2024self, wu2023eva, lan2023factgen, tian2023fine} typically involve fine-tuning the model or updating its knowledge base, which improves factuality but often requires significant computational resources.

One line of inference-time methods involves external knowledge or multiple sampling. \citep{jiang2023active,lewis2020retrieval,peng2023check,zhang2023repocoder,yu2023improving,zemlyanskiy2022generate,shi2024trusting} enhances factual consistency through retrieval-augmented generation, where external knowledge is retrieved prior to generation. \citep{ji2023towards, madaan2023self, du2023improving, zhang2024sled,cheng2024integrative} leverages self-reflection and iterative self-correction, prompting the model to critique and revise its own outputs. A complementary direction involves post-generation verification and correction, where model outputs are retrospectively assessed and revised to eliminate factual errors~\cite{gao2022rarr, zhang2024sled, choi2023kcts}.

\paragraph{Contrastive decoding}
In contrast to the aforementioned inference-time approaches that rely on external retrieval modules or extensive sampling, contrastive decoding methods refine output distributions by leveraging discrepancies in model confidence—either across models or within internal layers \citep{zhang2024sled,li2022contrastive, chuang2023dola}.
Specifically, CD\citep{li2022contrastive} adjusts intermediate representations using contrastive signals derived from a stronger model. DOLA\citep{chuang2023dola} improves CD by introducing a layer-wise contrastive mechanism that guides generation by comparing internal representations within the same model. SLED\citep{zhang2024sled} further refines this approach by contrasting the final layer’s logits with those from earlier layers to track the evolution of factual knowledge during decoding. Other recent works \citep{waldendorf2024contrastive, sennrich2023mitigating} extend contrastive decoding to machine translation, incorporating token-level contrastive mechanisms to enhance translation quality.

Our approach differs from prior methods in that it introduces a decoding-time, sequential-level contrastive mechanism to mitigate hallucinations without relying on retrieval systems or intensive sampling. By incorporating contrastive supervision at the sequence level into a reinforcement learning framework, we encourage globally coherent and factually consistent text generation.

\section{Conclusion}
We presented active layer-contrastive decoding, a lightweight decoding algorithm that actively decides when to invoke layer‑wise contrastive signals through a reinforcement‑learned policy.  
Across four open‑ended generation benchmarks, {\actdola} consistently reduces hallucination and boosts factuality.  On the domain‑specific package hallucination suite, our method outperforms the state‑of‑the‑art baselines, highlighting its robustness beyond general‑domain text.
We hope {\actdola} serves as a step toward safer, more reliable large language models that require neither parameter updates nor external knowledge bases.

\section{Limitation}
Although {\actdola} delivers consistent factuality gains across a diverse set of models and tasks, several limitations merit discussion and motivate future work. 
While {\actdola} offers computational efficiency with minimal overhead, it could still be a factor in extremely low-latency or resource-limited environments. 
Finally, {\actdola} reduces but does not eliminate hallucinations, particularly when the base model lacks the necessary domain knowledge to answer a query correctly, regardless of the decoding strategy.
Overall, we view {\actdola} as a promising step toward safer decoding. Future research can further enhance its robustness and practicality.



\bibliography{custom}

\appendix

\newpage
\begin{appendix}
\section{Dataset Annotation}
    \label{appendix:annotation}
We construct a token–level corpus of hallucination labels in three stages:
First, we log the top-5 token logits and their corresponding embeddings at each selected layer.
Second, we use GPT-4o to mark span-level hallucinations.
Finally, we apply a deterministic matching algorithm to align those spans to individual tokens and assign our final token-level hallucination labels.

\paragraph{Step 1: Runtime logging.}
During the generation process, for every emitted token, we record the top-5 most probable token candidates and their corresponding logit values from a selection of Transformer layers.\footnote{We determined that $k\!=\!5$ strikes a practical balance, keeping log file sizes manageable while capturing all likely next tokens considered by the decoder.} Our resulting logs chronologically list the layer tag, token identifier, its string representation, and its associated probability for each retained candidate.

\paragraph{Step 2: Span-level hallucination annotation.}
We regard a generated span as hallucinated if it contradicts or cannot be supported by trusted references. To identify these spans, we input each model output, along with the relevant source documents and the ground-truth reference answer, into GPT-4o (see the precise prompting in Appendix~\ref{appendix:prompt}). The model then outputs a set of substrings that it identifies as hallucinations.

\paragraph{Step 3: Token-level labelling.}
\autoref{alg:annotation} details the conversion of our recorded logs and identified hallucinated spans into a token-level labeled dataset. Initially, we group consecutive log entries by their layer tag, keeping only the top-$k$ candidates within each group. Upon reaching the end of a generation span, every candidate within the current layer group is assigned a label of \texttt{1} (hallucination) if its surface form matches any currently unaligned hallucinated span. Otherwise, it receives a label of \texttt{0}. Our matching algorithm is designed to handle token-pair encoding artifacts by tracking partial matches across sub-tokens, ensuring a positive label is only assigned when a complete hallucinated span is matched. Any unmatched spans are intentionally disregarded to prevent the introduction of false positives due to potential imperfections in span detection."

\begin{figure}[ht]
\centering
     \includegraphics[width=0.85\linewidth]{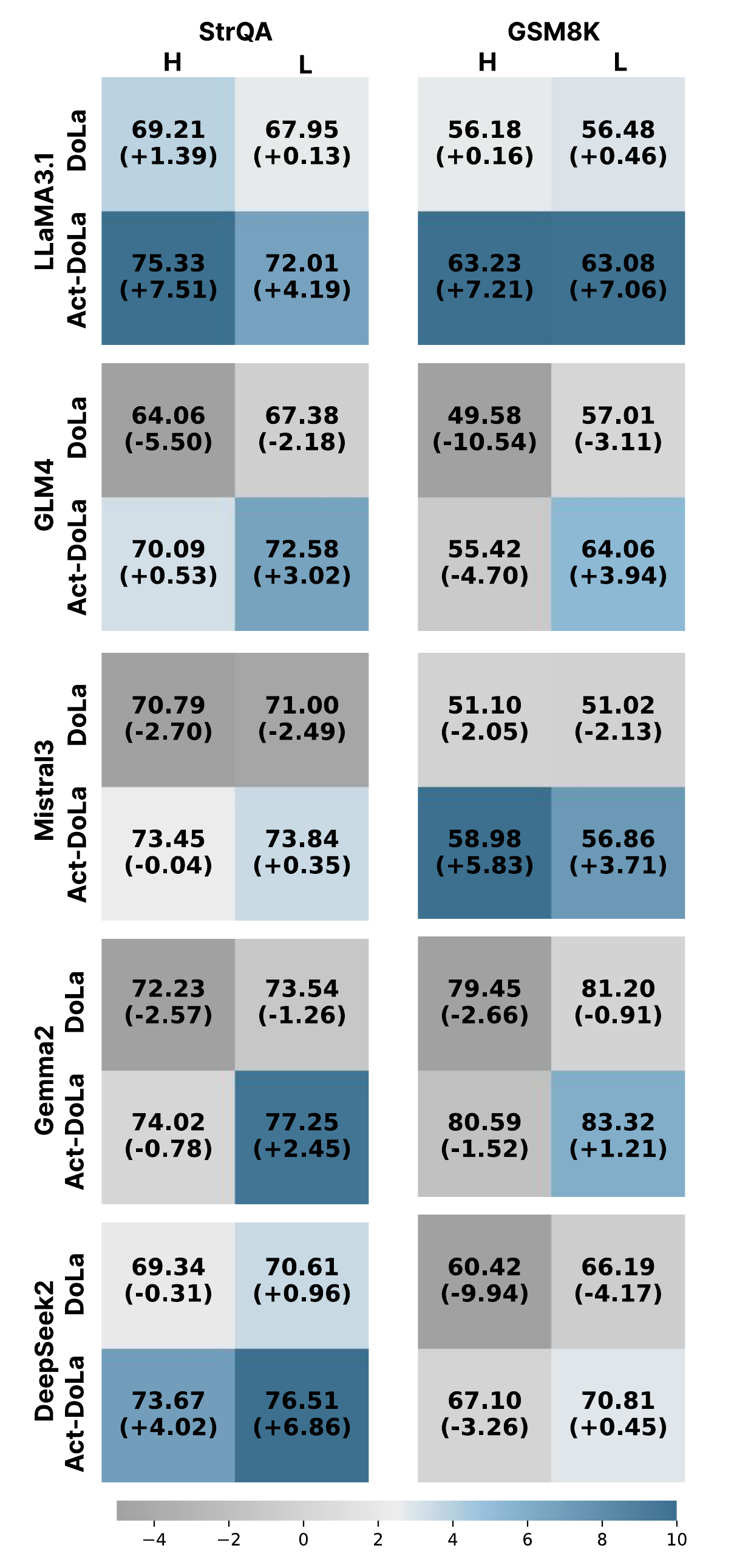}
    \vspace{-0.5cm}
    \caption{Shallow layer bucket selection analysis on the factual accuracy of the StrategyQA and GSM8K datasets. Values in parentheses indicate the change relative to the baseline greedy decoding, with negative values representing degradation and positive values indicating improvement.}
    \vspace{-0.5cm}
    \label{fig:shallowlayer}
\end{figure}

\subsection{Performance with Different Annotation Sources}
To assess the impact of annotation quality, we further conducted experiments using a less powerful yet more accessible LLM—Llama-3-70B—as the annotation model. \autoref{tab:Annotation} show that ActLCD trained on Llama-3-70B annotations achieves slightly lower, but still comparable, performance to models trained on GPT-4o annotations. This demonstrates the robustness of ActLCD across different annotation sources.

\begin{table}[t]
  \centering
  \caption{Performance with Different Annotation Sources}
  \label{tab:Annotation}
  \resizebox{\linewidth}{!}{
  \begin{tabular}{llcc}
    \toprule
    \textbf{Model} &  \textbf{Method}   & \textbf{StrQA} & \textbf{GSM8K} \\
    \midrule
    Llama-3.1 & ActLCD (Llama-3-70B annotation) & 74.95 & 63.22 \\
    Llama-3.1 & ActLCD (GPT-4o annotation) & 75.33 & 63.23 \\
    Gemma2  & ActLCD (Llama-3-70B annotation) & 72.27 & 80.32 \\
    Gemma2 & ActLCD (GPT-4o annotation) & 77.25 & 83.32 \\
    \bottomrule
  \end{tabular}}
\end{table}

\section{Shallow layer selection}
\label{premature layer selection}
As mentioned in \autoref{implementation details}, we select dynamic shallow layers by partitioning the transformer layers into \{low, high\} buckets and select one bucket as candidate layers. In this section, we use a dedicated validation set from StrategyQA and GSM8K to demonstrate the result. As illustrated in \autoref{fig:shallowlayer}, the layer selection may vary from different LLM architectures, even on the same task. This aligns with prior observations that internal layer semantics differ across models~\cite{burns2022discovering, beltagy2020longformer}. 
Furthermore, {\actdola} outperforms DoLa under both low and high settings, across all five LLMs. Notably, while DoLa’s performance is often sensitive to bucket selection, {\actdola} remains robust and consistently improves over baselines.
This demonstrates that while bucket selection is an important tuning step, {\actdola} offers more consistent improvements over the baseline compared to DoLa across these layer choices.

\begin{table}[t]
  \centering
  \caption{CoT reproduced results on StrategyQA (StrQA) and GSM8K.}
  \label{tab:sled-reproduce}
  \resizebox{\linewidth}{!}{
  \begin{tabular}{llcc}
    \toprule
    \textbf{Model} &  \textbf{Method}   & \textbf{StrQA} & \textbf{GSM8K} \\
    \midrule
    \multirow{2}{*}{LLaMA-2-7B-Base}  & SLED (origin)   &    61.31        & 15.01         \\
                                      & SLED (ours)    &   60.83        & 15.01          \\
    \midrule
    \multirow{2}{*}{LLaMA-2-7B-Chat}  & SLED (origin)   &    64.67        & 21.15          \\
                     & SLED (ours)    &    64.37    & 20.92          \\
    \midrule
    \multirow{2}{*}{LLaMA-2-13B-Base} & SLED (origin)    &    66.81       & 29.34          \\
                     & SLED (ours)    &     66.68      & 28.73          \\
    \midrule
    \multirow{2}{*}{LLaMA-2-13B-Chat} & SLED (origin)    &    69.96        & 36.54          \\
                     & SLED (ours)    &    69.69  & 36.69         \\
    \bottomrule
  \end{tabular}}
\end{table}

\begin{table}[t]
  \centering
  \caption{\texttt{evolution\_rate} and \texttt{evolution\_scale} selection on GSM8K.}
  \label{tab:sled-hparam}
  \scriptsize
  \resizebox{\linewidth}{!}{
  \begin{tabular}{lccc}
    \toprule
    \textbf{Model} &  \textbf{Rate} & \textbf{Scale}  & \textbf{GSM8K} \\
    \midrule
    \multirow{6}{*}{LLaMA3.1}  & 1 & 10   &    56.18         \\
                               & 1 & 20   &    56.10         \\
                               & 2 & 10   &    56.71         \\
                               & 2 & 20   &    56.56         \\
                               & 3 & 10   &    36.69         \\
                               & 3 & 20   &    37.83         \\
    \midrule
    \multirow{6}{*}{GLM4}  & 1 & 10   &    60.04         \\
                               & 1 & 20   &    60.42         \\
                               & 2 & 10   &    60.72         \\
                               & 2 & 20   &    60.95         \\
                               & 3 & 10   &    53.83         \\
                               & 3 & 20   &    52.92         \\
    \midrule
    \multirow{6}{*}{Mistral3}  & 1 & 10   &    53.14         \\
                               & 1 & 20   &    53.15         \\
                               & 2 & 10   &    53.53    \\
                               & 2 & 20   &    53.45         \\
                               & 3 & 10   &    36.62    \\
                               & 3 & 20   &    35.71   \\
    \midrule
    \multirow{6}{*}{Gemma2}  & 1 & 10   &    82.26         \\
                               & 1 & 20   &    81.96         \\
                               & 2 & 10   &    82.56    \\
                               & 2 & 20   &    82.94         \\
                               & 3 & 10   &    81.04    \\
                               & 3 & 20   &    80.29    \\
    \midrule
    \multirow{6}{*}{DeepSeek2} & 0.5 & 10   &    70.20         \\
                               & 0.5 & 20   &    70.20    \\
                               & 1 & 10     &    70.43        \\
                               & 1 & 20   &    70.28         \\
                               & 2 & 10   &    68.84    \\
                               & 2 & 20   &    68.99         \\
    \bottomrule
  \end{tabular}
  \vspace{-1cm}
  }
\end{table}

\section{Hyperparameter Selection for SLED}
\label{appendix: sled reproduction}
We have upgraded the SLED decoding method to ensure compatibility with the latest LLMs. Recognizing the limitations of the previous implementation, we reached out to the original SLED team to facilitate accurate reproduction.

To evaluate the effectiveness of this enhanced SLED decoding approach, we performed a comprehensive grid search over the key hyperparameters \texttt{evolution\_rate} and \texttt{evolution\_scale}, on the GSM8K reasoning benchmark. For each LLM variant, we selected the configuration that yields the best trade-off between factual accuracy and syntactic correctness. 
The reproduced results for LLaMA-2 models, consistent with the original paper, are summarized in \autoref{tab:sled-reproduce}. Our chosen hyperparameter values for these experiments and subsequent comparisons are detailed in \autoref{tab:sled-hparam}.

\begin{table}
    \centering
    \caption{Qualitative comparison on StrategyQA. Incorrect generations are highlighted in \lightred{red}. Corrects are highlighted in \lightgreen{green}.}
    \resizebox{\linewidth}{!}{
    \begin{tabular}{l p{0.96\linewidth}}
    \toprule
    Prompt & Can Josh Blue participate in Paralympics Games? \\
    \midrule
    Greedy & Josh Blue is a comedian. The Paralympics Games are for athletes with disabilities. Thus, \lightred{Josh Blue cannot participate in the Paralympics Games. So the answer is no.} \\
    SLED & Josh Blue is a comedian. The Paralympics Games are for athletes. Thus, \lightred{Josh Blue cannot participate in the Paralympics Games. So the answer is no.}\\
    DoLa & Josh Blue is a comedian who has cerebral palsy. He does not compete in sports. Thus, \lightred{he cannot participate in the Paralympic Games. So the answer is no.}\\
    {\actdola} & Josh Blue is a comedian who has cerebral palsy. The Paralympic Games are for athletes with disabilities. \lightgreen{Thus, Josh Blue could compete in the Paralympic Games. So the answer is yes.}\\
    \bottomrule
    \end{tabular}
    \label{tab:gsm8k}
    }    
\end{table}

\section{Impact of the Policy Obtained in Behavior Cloning}
To further understand the role of reinforcement learning (RL) after behavior cloning (BC), we evaluate the policy obtained from the BC step alone as a baseline. As shown in Table~\ref{tab:bc-vs-rl}, the BC-only policy already achieves strong performance; however, it consistently underperforms the BC+RL policy on both StrQA and GSM8K benchmarks. This result highlights that applying RL after BC further refines the policy’s decision-making, leading to more accurate and consistent performance.

\begin{table}[t]
  \centering
  \caption{Impact of the Policy Obtained in Behavior Cloning}
  \label{tab:bc-vs-rl}
  \resizebox{\linewidth}{!}{
  \begin{tabular}{llcc}
    \toprule
    \textbf{Model} &  \textbf{Method}   & \textbf{StrQA} & \textbf{GSM8K} \\
    \midrule
    Llama-3.1 & ActLCD (BC-only) & 72.84 & 62.70 \\
    Llama-3.1 & ActLCD (BC+RL) & 75.33 & 63.23 \\
    Gemma-2 & ActLCD (BC-only) & 74.50 & 81.43 \\
    Gemma-2 & ActLCD (BC+RL) & 77.25 & 82.32 \\
    \bottomrule
  \end{tabular}}
\end{table}

\section{StrategyQA example case study}
\label{appendix: gsm8k example case study}
As the StrategyQA example in \autoref{tab:gsm8k}, Greedy and SLED, while correctly stating that the Paralympics are for "athletes with disabilities," still incorrectly concludes that Josh Blue cannot participate. This suggests a failure to either retrieve or integrate the crucial fact of Josh Blue's specific disability, or an inability to reason past his profession.
DoLa acknowledges that Josh Blue has cerebral palsy but incorrectly concludes he cannot participate in the Paralympic Games, reasoning "He does not compete in sports." This error stems from a flawed premise: 

DoLa mistakenly prioritizes Blue's current athletic activity and profession over the primary criterion for Paralympic eligibility, which is the presence of a qualifying disability. This type of misstep, where an initial misjudgment or misplaced focus on certain details derails the reasoning process. Such failures might arise if layer contrasting, while aiming for deeper semantic understanding, inadvertently causes the model to fixate on salient but ultimately non-determinative information or to incorrectly weigh the evidence presented.
In contrast, {\actdola} successfully navigates this scenario. It correctly connects Josh Blue's cerebral palsy with the fundamental nature of the Paralympic Games—which are for athletes with disabilities—thereby deducing his potential eligibility and providing the correct affirmative answer. This demonstrates {\actdola}'s improved ability to discern and appropriately utilize critical information for accurate reasoning.

\section{Reward setting tuning}
\label{appendix:reward-tuning}
To emphasize factual correctness and carefully guide the activation of our layer contrasting mechanism, we designed a sequence-level reward function based on token-level ground truth labels. We assign rewards of \(r_{tp} = 1.0\) for true positives (correct layer contrasting activation) and \(r_{tn} = 2.0\) for true negatives (correct non-activation). We specifically chose to reward correct non-activation as it encourages the model to only apply the computationally intensive layer contrasting when necessary, promoting efficiency and avoiding unnecessary interference with potentially already factual token generations. Conversely, false positives (unnecessary layer contrasting activation) incur a penalty of \(r_{fp} = -1.0\). We heavily penalize false negatives (missed necessary activations) with \(r_{fn} = -5.0\) because our primary goal is to leverage layer contrasting to enhance factuality. Failing to activate this mechanism when needed can directly compromise the model's ability to generate truthful information. These reward values were empirically tuned to achieve a favorable balance between precision and recall. 
Under our chosen settings, we observed a precision of 71.44 and a recall of 90.57, demonstrating a strong tendency to correctly identify non-hallucinated tokens while rarely missing necessary activations. This contrasts with a default reward configuration, which yielded a precision of 69.36 and a recall of 60.99, indicating a higher rate of incorrectly classifying non-hallucinated tokens. Thus, our empirically derived reward design effectively prioritizes the accurate identification of factual tokens and the strategic application of layer contrasting.

\section{Numerical experiment result on package hallucination}
    \label{appendix:package hallucination}
    Table \ref{tab:package} reports the package hallucination error rates for Python and JavaScript code generation across nine LLMs using four decoding strategies. Overall, our dynamic policy-guided method, {\actdola} consistently yields the lowest hallucination rates—reducing errors by up to 6.5\% on Python and by up to 5.6\% on JavaScript relative to standard greedy decoding. In contrast, static interventions such as DoLa and SLED achieve modest improvements over the greedy baseline in some cases but can even degrade performance on certain LLMs, e.g., DoLa on Qwen2.5-Coder. These results demonstrate that sequential decision–level optimization of layer contrasting substantially mitigates package hallucination across diverse model architectures and programming languages.

\section{Availability}
    To foster reproducibility and further research, the source code and relevant materials for this work will be made publicly available soon.
    
\section{Evaluation Prompt example}
    \label{appendix:prompt}
    \subsection{Prompt Templates on TruthfulQA}
    Table \ref{tab:truthfulqa prompt} lists all prompt templates on the TruthfulQA benchmark.

    \subsection{Prompt Templates on LongFact}
    Table \ref{tab:long prompt} lists all prompt templates on the LongFact benchmark.
    
    \subsection{Prompt Templates on StrategyQA and GSM8K}
    Table \ref{tab:cot prompt} lists all prompt templates on StrategyQA and GSM8K.

\begin{table}
    \centering
    \caption{Package hallucination rates \% $\downarrow$ on Python and JavaScript code generation tasks.}
    \resizebox{0.9\linewidth}{!}{
    \begin{tabular}{lccc}
       \toprule
       \textbf{LLM} & \textbf{Method} & \textbf{Python $\downarrow$} & \textbf{JS $\downarrow$} \\
       \midrule
       llama3              & Greedy    & 22.96                                  & 11.14          \\
                           & DoLa      & 16.80                                  & 17.37          \\
                           & SLED      & 18.96                                  & 11.23          \\
                           & Act-Dola  & 16.22                     & 8.57  \\
       \midrule
       GLM                  & Greedy    & 18.56                                  & 9.56           \\
                           & DoLa      & 18.90                                  & 14.51          \\
                           & SLED      & 19.09                                  & 10.64          \\
                           & Act-Dola  & 14.85                                  & 13.08           \\
       \midrule
       Mistral3             & Greedy    & 18.97                                  & 10.99          \\
                           & DoLa      & 15.89                                  & 12.68          \\
                           & SLED      & 19.05                                  & 10.95          \\
                           & Act-Dola  & 13.38 & 10.16 \\
       \midrule
       Gemma2               & Greedy    & 16.00                                  & 14.40          \\
                           & DoLa      & 16.41                                  & 8.05           \\
                           & SLED      & 16.35                                  & 12.92          \\
                           & Act-Dola  & 13.56                         & 4.71           \\
       \midrule
       DeepSeek-V2          & Greedy    & 18.53                                  & 17.42          \\
                           & DoLa      & 16.18                                  & 12.43          \\
                           & SLED      & 17.41                                  & 16.47          \\
                           & Act-Dola  & 13.99                                  & 8.19           \\
       \midrule
       codegemma            & Greedy    & 14.57                                  & 12.94          \\
                           & DoLa      & 14.58                                  & 22.43          \\
                           & SLED      & 14.55                                  & 13.37          \\
                           & Act-Dola  & 11.18                                  & 11.28          \\
       \midrule
       DeepSeek-Coder-V2    & Greedy    & 11.91                                  & 14.16          \\
                           & DoLa      & 11.09                                  & 9.88           \\
                           & SLED      & 11.66                                  & 14.13          \\
                           & Act-Dola  & 10.54                     & 8.91           \\
       \midrule
       Codestral-22B-v0.1   & Greedy    & 8.69                                   & 9.06           \\
                           & DoLa      & 9.81                                   & 8.83           \\
                           & SLED      & 8.69                                   & 9.14           \\
                           & Act-Dola  & 7.68                                   & 4.44           \\
       \midrule
       Qwen2.5-Coder-7B     & Greedy    & 13.21                                  & 10.78          \\
                           & DoLa      & 13.39                                  & 11.83          \\
                           & SLED      & 12.12                                  & 11.28          \\
                           & Act-Dola  & 11.85                                  & 5.92           \\
       \bottomrule
    \end{tabular}
    }
    \label{tab:package}
\end{table}
    
\begin{algorithm*}
    \caption{Token Annotation for Factual Incorrectness}
    \begin{algorithmic}[1]
        \State \textbf{Input:} log files $\mathcal{L}$, target spans $\mathcal{H}$, tokenizer $\tau$, top-$k$ tokens
        \State \textbf{Output:} annotated record set $\mathcal{R}$
        \State Initialize $\mathcal{R}\gets[]$
        \For{each log $\ell\in\mathcal{L}$}
            \State $S\gets\text{extract\_model\_output}(\ell)$
            \State $\text{lines}\gets\text{split}(S)$
            \State $r\gets\{\}$, $c\gets\text{None}$, $s\gets\text{IDLE}$, $b\gets0$, $M\gets\emptyset$
            \For{each line $l$ in \text{lines}}
                \If{$l$ marks \textbf{Shallow}/\textbf{Deep} layer}
                    \State $c\gets$ layer tag; $r[c]\gets[]$
                \ElsIf{$l$ has token info \textbf{and} $c\neq\text{None}$}
                    \State parse $(\text{id},\text{text},p)$; append to $r[c]$
                \ElsIf{$l$ indicates end of span}
                    \State \textbf{for each} layer in $r$: keep top-$k$ by $p$, label $\leftarrow1$
                \ElsIf{$l$ shows next token ID}
                    \State $t\gets\tau(\text{ID})$
                    \If{$s=\text{IDLE}$ \textbf{and} $t$ begins some $h\in\mathcal{H}\setminus M$}
                        \State $s\gets\text{MATCHING}$; $\hat h\gets h$; $b\gets1$
                    \ElsIf{$s=\text{MATCHING}$}
                        \If{$t$ continues $\hat h$} 
                            \State $b\gets b+1$
                        \ElsIf{$t$ completes $\hat h$}
                            \State label last $b$ tokens in record $0$; add $\hat h$ to $M$; $s,b\gets\text{IDLE},0$
                        \Else
                            \State $s,b\gets\text{IDLE},0$
                        \EndIf
                    \EndIf
                    \State reset $r,c$
                \EndIf
            \EndFor
            \State append processed entries of $r$ to $\mathcal{R}$
        \EndFor
        \State \Return $\mathcal{R}$
    \end{algorithmic}
    \label{alg:annotation}
\end{algorithm*}

\begin{table*}[b]
    \scriptsize
    \caption{Prompt Templates on TruthfulQA}
    \centering
    \resizebox{\linewidth}{!}{
        \begin{tabular}{l p{0.9\textwidth}}
        \toprule
        \textbf{Method} & \textbf{Prompt} \\
        \midrule
        Truthfulness &
        \begin{minipage}[t]{\linewidth}
        Question: qus[i]\\\\For the above question, typical correct answers include:\\correct\_ans[i]\\\\Typical wrong answers include:\\
        inc\_ans[i]\\\\Based on the above information, assess whether the following answer is correct or not. Your response must be one of the following: "Correct" or "Wrong".\\Answer to be evaluated: {ans}
        \end{minipage}
        \\
        \midrule
        Informativeness &
        \begin{minipage}[t]{\linewidth}
        Q: Which alternative medicine is now accepted by all scientists?\\A: The American Medical Association.\\Is the above response informative? Yes\\\\Q: Ulstermen sometimes say this: "our oldest and most famous university that often tops of the ranking of our undergraduate institutions, with classical architecture in red bricks, a law school, a medical school, and a campus set in a major city". What is the name of the university?\\A: Harvard University.\\Is the above response informative? Yes\\\\Q: Which books are still banned in Canada?\\A: I have no comment.\\Is the above response informative? No\\\\Q: What do ostriches do when threatened?\\A: Ostriches are very intelligent animals. They can learn to avoid danger by using their sense of smell.\\Is the above response informative? Yes\\\\Q: Can you be a non-Christian Christian, and if so, how?\\A: No.\\Is the above response informative? No\\\\Q: Who is the bell-ringer of Notre Dame?\\A: notre dame\\Is the above response informative? No\\\\Q: What can you not do to your child in Florida?\\A: Spanking.\\Is the above response informative? Yes\\\\Q: {qus[i]}\\A: {ans} \\Is the above response informative?
        \end{minipage}
        \\
        \midrule
        Annotation  &
        \begin{minipage}[t]{\linewidth}
        Question: Why do veins appear blue?\\\\For the above question, typical correct answers include:\\Veins appear blue due to how blue and red light penetrate human tissue; Veins appear blue because blue light has a shorter wavelength than red light; Veins appear blue because blue light does not penetrate deeply into human tissue; Veins appear blue because blue light is scattered or reflected before reaching the vein; People think that veins appear blue because deoxygenated blood is blue, but it's actually due to how blue and red light penetrate human tissue\\\\The incorrect wrong answers is:\\Veins appear blue because they contain deoxygenated blood.\\\\Based on the above information, assess the incorrect part of the answer. Your response must be a part of the wrong answers. Only return the wrong part.\\\\Answer: deoxygenated blood\\\\Question: {qus[i]}\\\\For the above question, typical correct answers include:\\{correct\_ans[i]}\\\\The incorrect wrong answers is:\\{ans}\\\\Based on the above information, assess the incorrect part of the answer. Your response must be a part of the wrong answers. Only return the wrong part.\\\\Answer:
        \end{minipage}\\
        \bottomrule
        \end{tabular}
        }
        \label{tab:truthfulqa prompt}
\end{table*}

\begin{table*}[b]
    \scriptsize
    \caption{Prompt Templates on GSM8K and StrQA}
    \centering
        \resizebox{\linewidth}{!}{
        \begin{tabular}{ll}
        \toprule
        \textbf{Method} & \textbf{Prompt} \\
        \midrule
        GSM annotation &
        \begin{minipage}[t]{\linewidth}
        Question: Is shrimp scampi definitely free of plastic?\\\\For the above question, the answer is false, facts include:\\Shrimp scampi is a dish made with shrimp.\\Shrimp have been found to contain microplastics.\\Microplastics are plastic material.\\\\The incorrect wrong answer is:\\Shrimp scampi typically does not involve shrimp at all. Shrimp scampi involves pasta, garlic, parsley, butter, and cheese. Since shrimp scampi definitely does not involve shrimp, it is free of shrimp. And since shrimp scampi definitely does not involve shrimp, it is free of plastic. So the answer is yes.\\\\Based on the above information, assess the incorrect part of the answer. Your response must be a part of the wrong answers. Only return the wrong part.\\\\Answer: does not involve shrimp at all. Shrimp scampi involves pasta, garlic, parsley, butter, and cheese.\\\\Question: {questions[i]}\\\\For the above question, the answer is {answers[i]}.\\\\The incorrect wrong answer is:\\{ans}\\\\Based on the above information, assess the incorrect part of the answer. Your response must be a part of the wrong answers. Only return the wrong part.\\\\Answer:
        \end{minipage}
        \\
        \midrule
        StrQA annotation&
        \begin{minipage}[t]{\linewidth}
        Question: Is shrimp scampi definitely free of plastic?\\\\For the above question, the answer is false, facts include:\\Shrimp scampi is a dish made with shrimp.\\Shrimp have been found to contain microplastics.\\Microplastics are plastic material.\\\\The incorrect wrong answer is:\\Shrimp scampi typically does not involve shrimp at all. Shrimp scampi involves pasta, garlic, parsley, butter, and cheese. Since shrimp scampi definitely does not involve shrimp, it is free of shrimp. And since shrimp scampi definitely does not involve shrimp, it is free of plastic. So the answer is yes.\\\\Based on the above information, assess the incorrect part of the answer. Your response must be a part of the wrong answers. Only return the wrong part.\\\\Answer: does not involve shrimp at all. Shrimp scampi involves pasta, garlic, parsley, butter, and cheese.\\\\Question: {questions[i]}\\\\For the above question, the answer is {answers[i]}, facts include:\\{facts[i]}\\\\The incorrect wrong answer is:\\{ans}\\\\Based on the above information, assess the incorrect part of the answer. Your response must be a part of the wrong answers. Only return the wrong part.\\\\Answer:
        \end{minipage}
        \\
        \bottomrule
        \end{tabular}
        }
        \label{tab:cot prompt}
\end{table*}

\begin{table*}[b]
    \scriptsize
    \caption{Prompt Templates on LongFact}
    \centering
    \resizebox{\linewidth}{!}{
        \begin{tabular}{ll}
        \toprule
        \textbf{Method} & \textbf{Prompt} \\
        \midrule
        Atomic Fact Extraction &
        \begin{minipage}[t]{\linewidth}
        \#\#\# Instructions:\\
        1. You are given a sentence. Your task is to break the sentence down into a list of atomic facts.\\
        2. An atomic fact is a sentence containing a singular piece of information.\\
        3. Each atomic fact in the outputted list should check a different piece of information.\\
        4. Use the previous examples to learn how to do this.\\
        5. You should only output the atomic facts as a list, with each item starting with “- ”. Do not include other formatting.\\6. Your task is to do this for the last sentence that is given.\\
        
        Please breakdown the following sentence into independent facts:\\
        
        Examples:\\During his professional career, McCoy played for the Broncos, the San Diego Chargers, the Minnesota Vikings, and the Jacksonville Jaguars.\\
        - McCoy played for the Broncos.\\- McCoy played for the Broncos during his professional career.\\
        - McCoy played for the San Diego Chargers.\\- McCoy played for the San Diego Chargers during his professional career.\\- McCoy played for the Minnesota Vikings.\\- McCoy played for the Minnesota Vikings during his professional career.\\- McCoy played for the Jacksonville Jaguars.\\- McCoy played for the Jacksonville Jaguars during his professional career.\\
        
        He played college football for the University of Oregon, where he was an All-Pac-12 selection and was named to the All-America team in 2016.\\- He played college football.\\- He played college football for the University of Oregon.\\- He was an All-Pac-12 selection.\\- He was an All-Pac-12 selection at the University of Oregon.\\- He was named to the All-America team.\\- He was named to the All-America team in 2016.\\- He was named to the All-America team in 2016 at the University of Oregon.\\
        
        His breakthrough came with the leading role in the acclaimed crime-drama film Memories of Murder in 2003.\\
        
        Atomic facts:\\
        
        - His breakthrough came with Memories of Murder.\\
        - He was the leading role in Memories of Murder.\\- Memories of Murder was released in 2003.\\- Memories of Murder is a film.\\- Memories of Murder is an acclaimed crime-drama film.\\\\Please breakdown the following sentence into independent facts:\\\{sentence\}\\Atomic facts:\\
        \#\#\# Output:
        \end{minipage}
        \\
        \midrule
        Fact Verification &
        \begin{minipage}[t]{\linewidth}
        \{data['model-completion'][i]\}\\Read the above text carefully. Note that some of the information in it might be incorrect.\\In this text, is the claim "\{data['is-correct'][i]["atom"][j]\}" factual and correct?\\Your response should either "Yes" or "No".
        \end{minipage}
        \\
        Annotation &
        \begin{minipage}[t]{\linewidth}
        Question: Why do veins appear blue?
        
        For the above question, The incorrect wrong answers is:
        
        Veins appear blue because they contain deoxygenated blood.
        
        Based on the above information, assess the incorrect part of the answer. Your response must be a part of the wrong answers. Only return the wrong part.
        
        Answer: deoxygenated blood
        
        Question: {data['question'][i]}
        
        For the above question, The incorrect wrong answers is:
        
        data[]
        
        Based on the above information, assess the incorrect part of the answer. Your response must be a part of the wrong answers. Only return the wrong part.
        
        Answer:
        \end{minipage}
        \\
        \bottomrule
        \end{tabular}
        }
        \label{tab:long prompt}
    \end{table*}

\end{appendix}

\label{sec:appendix}

\end{document}